\documentclass{article}

\usepackage{microtype}
\usepackage{graphicx}
\usepackage{subfigure}
\usepackage{booktabs} 

\usepackage{hyperref}

\usepackage{listings} 
\usepackage[accepted]{icml2023_deployableGenAI}

\usepackage{amsmath}
\usepackage{amssymb}
\usepackage{mathtools}
\usepackage{amsthm}

\usepackage[capitalize,noabbrev]{cleveref}

\theoremstyle{plain}

\theoremstyle{definition}

\theoremstyle{remark}

\usepackage[textsize=tiny]{todonotes}

\icmltitlerunning{DisasterResponseGPT: LLMs for Accelerated Plan of Action Development in Disaster Response Scenarios}

\begin{document}

\twocolumn[
\icmltitle{DisasterResponseGPT: Large Language Models for Accelerated Plan of Action Development in Disaster Response Scenarios}




\begin{icmlauthorlist}
\icmlauthor{Vinicius G. Goecks}{arl}
\icmlauthor{Nicholas R. Waytowich}{arl}
\end{icmlauthorlist}

\icmlaffiliation{arl}{DEVCOM Army Research Laboratory, Aberdeen Proving Ground, Maryland, USA}

\icmlcorrespondingauthor{Vinicius G. Goecks}{vinicius.goecks@gmail.com}
\icmlcorrespondingauthor{Nicholas R. Waytowich}{nicholas.r.waytowich.civ@army.mil}

\icmlkeywords{Machine Learning, ICML}

\vskip 0.3in
]



\printAffiliationsAndNotice{}  

\begin{abstract}
The development of plans of action in disaster response scenarios is a time-consuming process. Large Language Models (LLMs) offer a powerful solution to expedite this process through in-context learning. This study presents DisasterResponseGPT, an algorithm that leverages LLMs to generate valid plans of action quickly by incorporating disaster response and planning guidelines in the initial prompt. In DisasterResponseGPT, users input the scenario description and receive a plan of action as output. The proposed method generates multiple plans within seconds, which can be further refined following the user's feedback. Preliminary results indicate that the plans of action developed by DisasterResponseGPT are comparable to human-generated ones while offering greater ease of modification in real-time. This approach has the potential to revolutionize disaster response operations by enabling rapid updates and adjustments during the plan's execution.
\end{abstract}

\section{Introduction}

Disaster response operations are complex and dynamic, requiring users to make critical decisions under time-sensitive and high-pressure conditions \cite{Rennemo2014ATS,Jayawardene2021TheRO,Uhr2018AnES}. One of the key elements in the decision-making process is the development of plans of action, which represent alternative steps to achieve objectives while taking into account various operational constraints. Traditionally, plan of action development is a time-consuming process that relies heavily on the experience and expertise of the disaster response personnel. With lives at risk, a more efficient approach to develop plans of action is needed \cite{kovel2000modeling,alsubaie2013platform,Rennemo2014ATS,SUN2021107213}.

Large Language Models (LLMs) have emerged as a powerful tool for natural language processing \cite{devlin2018bert,raffel2020exploring,brown2020language}, offering potential applications in various domains, including disaster response \cite{ningsih2021disaster,zhou2022victimfinder,ghosh2022gnom}. LLMs can process vast amounts of data and generate human-like text based on given prompts \cite{scao2021many,min2022rethinking,xie2021explanation}. This research paper explores the application of LLMs to accelerate the development of plans of action for disaster response operations through in-context learning.

This work presents DisasterResponseGPT, a framework that leverages LLMs to quickly develop valid plans of action by incorporating disaster response and planning guidelines in the initial system prompt. Users provide a description of the scenario and main goals and receive a plan of action as output. The proposed method generates multiple plans of action within seconds, significantly reducing the overall development time. This enables a faster and more agile decision-making process, allowing users to quickly iterate on plans of action, adapt to changes in the scenario and rapidly start the rescue operations.

\begin{figure}[ht]
    \centering
    \includegraphics[width=0.9\columnwidth]{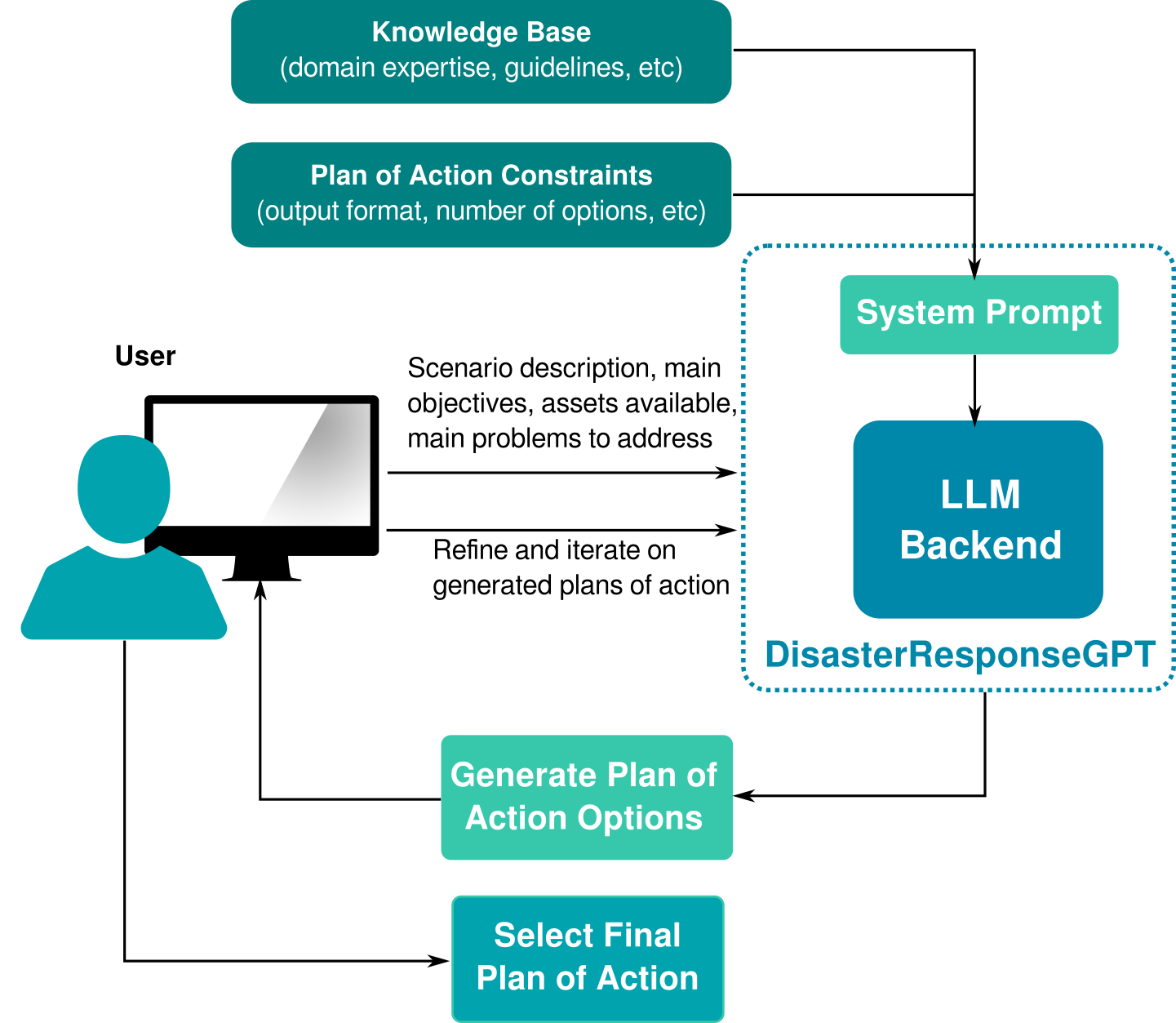}
    \caption{Overview of DisasterResponseGPT. It consists of an LLM initially prompted with a knowledge base and additional constraints for in-context learning. A human user supplies information about the disaster response scenario and DisasterResponseGPT generates options for plans of action, iterating with the users via natural language until they decide on the final plan of action.}
    \label{fig:diagram}
\end{figure}

Our main contributions are:
\begin{itemize}
    \item DisasterResponseGPT, a framework that leverages LLMs to develop valid plans of action for disaster response scenarios quickly;
    \item Experiments with state-of-the-art LLMs indicating that LLM-generated plans of action are comparable in quality to those developed by humans; and
    \item A support tool for authorities to generate plans of action with greater ease of modification in real-time, allowing for rapid adjustments during rescue operations.
\end{itemize}

\section{Methods}

In this research, we utilized the in-context learning capabilities of large language models (LLMs) to develop a chatbot assistant, DisasterResponseGPT, that can efficiently generate plans of action for disaster response operations. To achieve this, we incorporated information from the Federal Emergency Management Agency (FEMA) and examples of desired outputs in the initial prompt to bias the LLM towards generating plans of action (Appendix \ref{appendix:system_prompt}).

As illustrated in Figure \ref{fig:diagram}, DisasterResponseGPT assistant interacts with users via text, requesting them to provide a scenario description, main objectives, available assets, and main problems to be addressed. Based on the input, DisasterResponseGPT generates three plans of action that outline the main objective and what is critical to complete it, the main and auxiliary operations of the plan, the end states for the operation, and how each plan of action is feasible, acceptable, and suitable for the scenario.

Users are presented with these three options, allowing them to choose their preferred plan of action and refine it according to their needs by providing suggestions through text. DisasterResponseGPT processes this feedback and refines the selected plan of action accordingly. Depending on the specific LLM used in the backend, DisasterResponseGPT may also attempt to generate a visual sketch of the planned operations or instructions to generate one. In this research work, we compared results with three different LLMs in the backend: GPT-3.5 \cite{ouyang2022training} and GPT-4 \cite{openai2023gpt4,bubeck2023sparks} --- models ``\textit{gpt-3.5-turbo}'' and ``\textit{gpt-4}'', respectively, via OpenAI's API \footnote{OpenAI Chat Completion documentation: \url{https://platform.openai.com/docs/guides/chat}.} --- and Bard Experiment \cite{anil2023palm} via the browser interface\footnote{Google Bard Experiment: \url{https://bard.google.com/}.}. 

The plans of action generated by DisasterResponseGPT are produced in a matter of seconds, and taking into account the entire user interaction time, a final plan of action can be achieved within a few minutes. This streamlined process demonstrates the potential of DisasterResponseGPT to revolutionize plan of action development for disaster response operations, allowing for rapid updates and adjustments during the plan's execution to address discrepancies in the planning phase.

\section{Results and Discussion}

All experiments presented in this paper simulate the same fictional disaster response scenario:
\begin{quote}
    On April 16th, 2023, an earthquake of significant magnitude struck a small city nestled within a valley in California. The seismic event triggered a catastrophic landslide that blocked the main access road to the area, effectively cutting off the city from external assistance. Additionally, the landslide wreaked havoc across the adjacent residential zones, with several houses reported destroyed. While rescue teams are striving to reach potential survivors, the situation remains critical due to visible fractures on the hill that indicate a high risk of subsequent landslides. \textbf{Your main objective is to restore accessibility to the city by clearing and securing the blocked roadway}.
\end{quote}

\begin{figure}[ht]
    \centering
    \includegraphics[width=0.9\columnwidth]{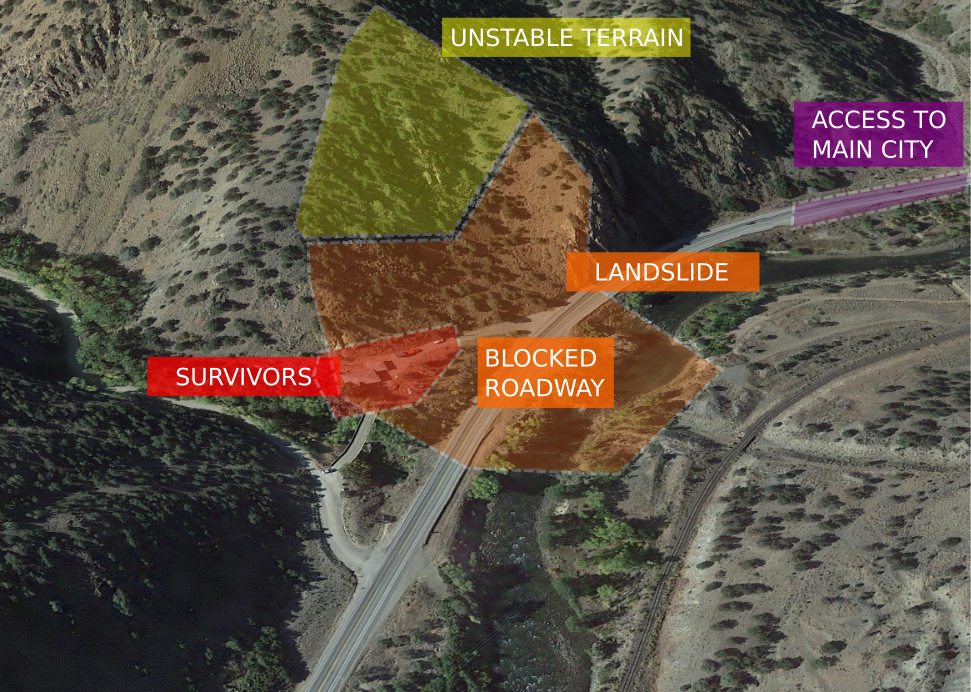}
    \caption{Visualization of the scenario investigated in this work. An earthquake-triggered landslide destroys a residential area and blocks the main access to the city. There might be survivors in the affected area and the roadway needs to be cleared so the emergency response teams can also reach the city that was also affected by the earthquake. The unstable terrain on top of the hill might also lead to new landslides.}
    \label{fig:scenario}
\end{figure}

In this scenario, as shown in Figure \ref{fig:scenario}, the available assets for the disaster response operation and the main problems to be addressed are as follows:
\begin{itemize}
    \item \textbf{Available assets}: one emergency response team equipped with heavy-duty excavation and construction equipment, two disaster response units with search and rescue dogs, one medical team for immediate on-site treatment, and one geotechnical team.
    \item \textbf{Main problems to be addressed}: potential for ongoing geological instability and the blocked access road, which severely hampers rescue efforts and supply routes. Lives are at risk from people potentially trapped in the destroyed houses before and after the road blockage and from the threat of further landslides.
\end{itemize}

The experiment begins with the user starting the DisasterResponseGPT chat interface. In the background, DisasterResponseGPT is prompted at the system level with the required information to generate plans of action for disaster response scenarios, as shown in Appendix \ref{appendix:system_prompt}. DisasterResponseGPT then asks the user for the scenario description, main objectives, available assets and main problems to be addressed.
Next, a message automatically aggregating this information is prepared and sent to DisasterResponseGPT, which in turn replies with three developed plans of action describing the main objective and what is critical to complete it, main and auxiliary operations of the plan (including tasks, purpose, and which assets are performing each task), the end states for the operation, and how each plan of action is feasible, acceptable, and suitable for the scenario.
The user interacts with DisasterResponseGPT via language to select a refine a plan of action until the user is satisfied with the final output.
We compared the plan of action generated by a human against the ones generated by DisasterResponseGPT with three different LLM as backend: GPT-3.5 and GPT-4 by OpenAI, and Bard by Google.
The full conversation transcripts for each baseline are available in Appendix \ref{appendix:transcripts}.

\begin{table*}[!ht]
\tiny
\caption{Final Plans of Action generated by DisasterResponseGPT compared to a human-generated plan.}
\label{table:experiments}
\begin{tabular}{@{}p{0.06\textwidth}p{0.2\textwidth}p{0.2\textwidth}p{0.2\textwidth}p{0.2\textwidth}@{}}
\toprule
\multicolumn{1}{c}{\textbf{Statement Item}} & \multicolumn{1}{c}{\textbf{Human}} & \multicolumn{1}{c}{\textbf{GPT-4 Backend}} & \multicolumn{1}{c}{\textbf{GPT-3.5 Backend}} & \multicolumn{1}{c}{\textbf{Bard Backend}} \\ \midrule
Objective & Restore accessibility to the city by clearing and securing the blocked roadway. & Restore accessibility to the city by clearing and securing the blocked roadway. & Restore accessibility by clearing and securing the blocked roadway. & Restore accessibility to the city by clearing and securing the blocked roadway. \\ \midrule
Critical to this objective & Clearing and securing the blocked roadway. & Stabilizing the hill to prevent further landslides while clearing debris from the access road. & To prioritize the removal of debris from the roadway to restore accessibility while minimizing the risk to the response team by utilizing protective measures. & The safety of the rescue workers and the potential survivors. The ability to quickly and efficiently clear the roadway of debris. The ability to secure the roadway to prevent further landslides. \\ \midrule
Main Operations & Task 1: clear and secure the blocked roadway. Purpose 1: re-establish accessibility to the city to allow the transport of supplies and rescue teams to other affected areas. Assets performing task: one emergency response team equipped with heavy-duty excavation and construction equipment. & Task 1: Stabilize the hill to prevent further landslides. Purpose 1: Enable safe debris removal from the access road. Assets performing task: Geotechnical team. Task 2: Clear and secure the access road. Purpose 2: Restore access to the city. Assets performing task: Emergency response team equipped with heavy-duty excavation and construction equipment, disaster response units with search and rescue dogs. & Task 1: Remove debris from the roadway using excavation machines and dump trucks to clear access to the city. Purpose 1: To restore the accessibility. Assets performing task: Emergency response team equipped with heavy-duty excavation and construction equipment. Task 2: Stabilize the slope by dewatering to mitigate the risk of ongoing geological instability. Purpose 2: To prevent subsequent landslides. Assets performing task: Geotechnical team and emergency response team equipped with heavy-duty excavation and construction equipment. & Task 1: Assess the situation and develop a plan of action. Purpose 1: To determine the best way to clear the roadway and secure it against further landslides. Assets performing task: The geotechnical team. Task 2: Clear the roadway of debris. Purpose 2: To remove any obstacles that are preventing access to the city. Assets performing task: The emergency response team with heavy-duty excavation and construction equipment. Task 3: Secure the roadway against further landslides. Purpose 3: To prevent the roadway from collapsing again and blocking access to the city. Assets performing task: The geotechnical team. \\ \midrule
Auxiliary Operations & Task 2: assess risk of subsequent landslides. Purpose 2: ensure that roadway-clearing team can perform their task safely. Assets performing task: one geotechnical team. Task 3: search for survivors in the residential zone affected by the first landslide. Purpose 3: rescue survivors from initial landslide. Assets performing task: one disaster response units with search and rescue dogs and one medical team for immediate on-site treatment. Task 4: search for survivors in the main city zone affected by the earthquake. Purpose 4: rescue survivors from earthquake accessing the city via alternative route. Assets performing task: one disaster response units with search and rescue dogs. & Task 3: Locate and rescue trapped individuals in destroyed houses and adjacent residential zones affected by the landslide. Purpose 3: Save as many lives as possible. Assets performing task: Disaster response units with search and rescue dogs, medical team for immediate on-site treatment. & Task 3: Conduct search and rescue operations in all destroyed houses before the road blockage to mitigate loss of life from people potentially trapped. Purpose 3: To identify and save potential survivors. Assets performing task: One disaster response unit with search and rescue dogs and one medical team for immediate on-site treatment. Task 4: Conduct search and rescue operations in the destroyed houses after removing debris to ensure no one is trapped or injured since the road blockage. Purpose 4: To identify and save potential survivors. Assets performing Task: One disaster response unit with search and rescue dogs and one medical team for immediate on-site treatment. & Task 4: Search for and rescue potential survivors. Purpose 4: To locate and rescue any people who may be trapped in the destroyed houses. Assets performing task: The disaster response units with search and rescue dogs. Task 5: Provide medical care to the injured. Purpose 5: To provide immediate medical care to the injured survivors. Assets performing task: The medical team. \\ \midrule
End states & Assets: all assets completed their task safely with no casualties. Victims: all victims were removed from the debris and given medical treatment. Civilians: evacuated to safe areas and temporary shelters. Terrain: roadway cleared allowing the transport of supplies and rescue teams to other affected areas. & Assets: All teams and equipment are safely withdrawn from the area after road accessibility has been restored. Victims: Rescued individuals receive immediate on-site treatment from the medical team and are evacuated. Civilians: Civilians are informed about the stable access road and the possibility of evacuation or receiving emergency supplies. Terrain: The access road is clear and secured, and potential landslide areas on the hill are stabilized. & Assets: Emergency response team equipped with heavy-duty excavation and construction equipment and geotechnical team have cleared the access to the city and mitigated risk of ongoing geological instability. Victims: Lives of people potentially trapped in destroyed homes before and after the road blockage have been saved. Civilians: Access to the city has been restored. Terrain: Slope stability has been stabilized. & The roadway is clear and secure. The city is accessible to external assistance. The potential survivors have been rescued and are receiving medical care. \\ \midrule
\end{tabular}
\end{table*}

\begin{figure}[!ht]
    \centering
    \includegraphics[width=0.9\columnwidth]{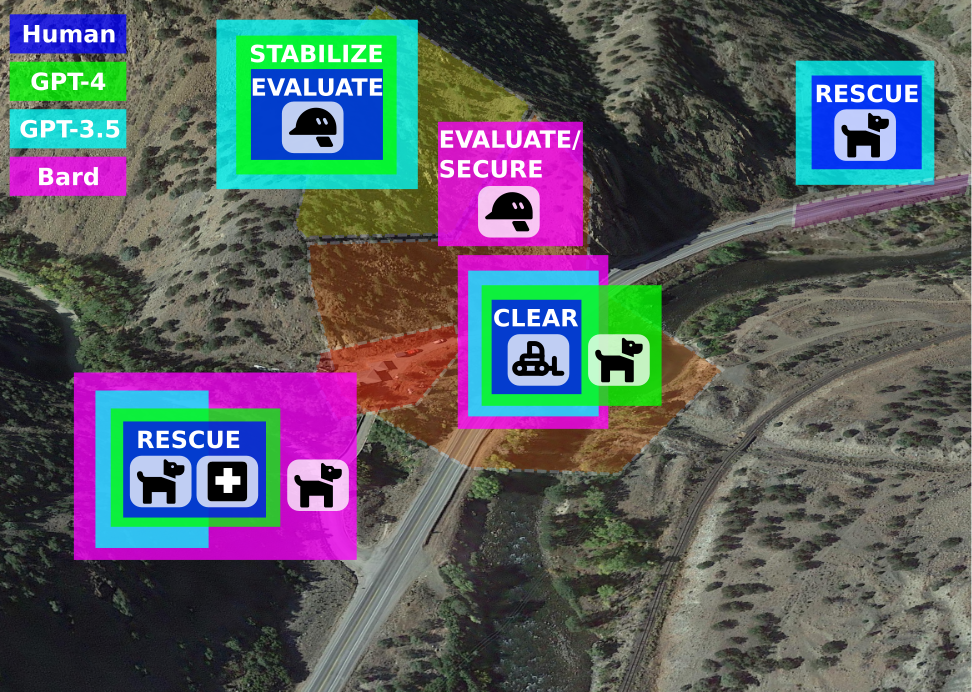}
    \caption{Overview of the task allocation proposed by each model and the human baseline. A symbol next to an affected area means that an asset was assigned to address the problem at that location. The hardhat symbol represents the geotechnical team, the bulldozer represents the emergency response team equipped with heavy-duty excavation and construction equipment, the dog represents the disaster response unit with search and rescue dogs, and the cross represents the medical team for immediate on-site treatment.}
    \label{fig:scenario_plans}
\end{figure}

Table \ref{table:experiments} compares the plan of action generated by a human to the final plans of action developed by DisasterResponseGPT using GPT-3.5, GPT-4, and Bard models as the LLM in the backend, here addressed as DisasterResponseGPT-3.5, DisasterResponseGPT-4, and DisasterResponseGPT-Bard, respectively, for conciseness.
For visualization, Figure \ref{fig:scenario_plans} presents a summary of the tasks performed by each asset for each final plan of action presented in Table \ref{table:experiments}.
Overall, all baselines addressed the main problems described in the scenario and were able to successfully output a plan of action in mostly the same format as defined in the system prompt, although GPT-3.5 and GPT-4 backends complied better with the requested format when compared to Bard.

DisasterResponseGPT-3.5 closest matched the plan generated by the human, including being the only model to allocate assets to the rescue of survivors after the road blockage, as the human-generated plan did.
DisasterResponseGPT-3.5 and DisasterResponseGPT-Bard, differently from the human and DisasterResponseGPT-4, proposed to use the same asset in multiple tasks, creating the perception that tasks were to be performed sequentially.
For example, DisasterResponseGPT-Bard first allocated the geotechnical team to assess the situation and then to secure the roadway against further landslides.

Sketches offer a tangible representation of plans of action, making them more accessible and easily comprehensible to the teams performing the operations.
Unfortunately, all LLMs investigated in this work were unable to automatically generate sketches when asked to present a visualization of the final plan of action.
At best, users were given instructions on how to generate one themselves.
We expect that future multimodal versions of LLMs with image manipulation capabilities will be able to overcome this limitation.

\section{Conclusions}

In this paper we present DisasterResponseGPT, a highly effective method for accelerating the development of plans of action for disaster response operations by leveraging the Large Language Models (LLMs).
A notable advantage of the DisasterResponseGPT method is the ease with which users can directly interact with and refine plans of action in real-time. This interactive capability allows for the exploration of a wide range of options during planning, increasing the likelihood of identifying the most effective plan of action. Furthermore, DisasterResponseGPT integrates authority guidelines and expert knowledge effectively, with the potential to expand its knowledge database easily as new information becomes available.

The rapid generation of plans of action by DisasterResponseGPT holds significant potential for faster rescue operations, which is essential for saving lives. The use of either GPT-3.5, GPT-4, or Bard in the backend ensures that the generated plans of action are of high quality and comparable to those developed by a human. Other LLMs of similar size should be able to achieve similar performance.

While the DisasterResponseGPT method is capable of generating plans of action, the automatic generation of usable sketches remains a challenge. Nevertheless, the DisasterResponseGPT approach is a groundbreaking development with the potential to revolutionize disaster response planning by enabling rapid updates and adjustments during rescue operations, addressing discrepancies in the planning phase.

\subsection{Limitations}

In this research, our investigation of LLMs is constrained by two primary limitations. First, the LLMs investigated are currently unable to process image-based inputs. Consequently, we have explored providing the scenario only in plain text descriptions. Second, the context size of the LLMs under investigation is restricted to a maximum of 4096 and 8192 tokens. This constraint directly impacts the number of plans of action that can be generated, the amount of interaction between the user and DisasterResponseGPT, as well as the potential inclusion of supplementary outputs such as sketches. We anticipate revisiting and addressing these limitations as multimodal LLMs and larger context window capabilities become more prevalent in the future.

\section*{Acknowledgements}

This research was sponsored by the Army Research Laboratory and was accomplished under Cooperative Agreement Number W911NF-23-2-0072. The views and conclusions contained in this document are those of the authors and should not be interpreted as representing the official policies, either expressed or implied, of the Army Research Laboratory or the U.S. Government. The U.S. Government is authorized to reproduce and distribute reprints for Government purposes notwithstanding any copyright notation herein.

\bibliography{refs}

\begin{thebibliography}{19}
\providecommand{\natexlab}[1]{#1}
\providecommand{\url}[1]{\texttt{#1}}
\expandafter\ifx\csname urlstyle\endcsname\relax
  \providecommand{\doi}[1]{doi: #1}\else
  \providecommand{\doi}{doi: \begingroup \urlstyle{rm}\Url}\fi

\bibitem[Alsubaie et~al.(2013)Alsubaie, Di~Pietro, Marti, Kini, Lin, Palmieri,
  and Tofani]{alsubaie2013platform}
Alsubaie, A., Di~Pietro, A., Marti, J., Kini, P., Lin, T.~F., Palmieri, S., and
  Tofani, A.
\newblock A platform for disaster response planning with interdependency
  simulation functionality.
\newblock In \emph{Critical Infrastructure Protection VII: 7th IFIP WG 11.10
  International Conference, ICCIP 2013, Washington, DC, USA, March 18-20, 2013,
  Revised Selected Papers 7}, pp.\  183--197. Springer, 2013.

\bibitem[Anil et~al.(2023)Anil, Dai, Firat, Johnson, Lepikhin, Passos, Shakeri,
  Taropa, Bailey, Chen, et~al.]{anil2023palm}
Anil, R., Dai, A.~M., Firat, O., Johnson, M., Lepikhin, D., Passos, A.,
  Shakeri, S., Taropa, E., Bailey, P., Chen, Z., et~al.
\newblock {PaLM 2} technical report.
\newblock \emph{arXiv preprint arXiv:2305.10403}, 2023.

\bibitem[Brown et~al.(2020)Brown, Mann, Ryder, Subbiah, Kaplan, Dhariwal,
  Neelakantan, Shyam, Sastry, Askell, et~al.]{brown2020language}
Brown, T., Mann, B., Ryder, N., Subbiah, M., Kaplan, J.~D., Dhariwal, P.,
  Neelakantan, A., Shyam, P., Sastry, G., Askell, A., et~al.
\newblock Language models are few-shot learners.
\newblock \emph{Advances in neural information processing systems},
  33:\penalty0 1877--1901, 2020.

\bibitem[Bubeck et~al.(2023)Bubeck, Chandrasekaran, Eldan, Gehrke, Horvitz,
  Kamar, Lee, Lee, Li, Lundberg, et~al.]{bubeck2023sparks}
Bubeck, S., Chandrasekaran, V., Eldan, R., Gehrke, J., Horvitz, E., Kamar, E.,
  Lee, P., Lee, Y.~T., Li, Y., Lundberg, S., et~al.
\newblock Sparks of artificial general intelligence: Early experiments with
  {GPT-4}.
\newblock \emph{arXiv preprint arXiv:2303.12712}, 2023.

\bibitem[Devlin et~al.(2018)Devlin, Chang, Lee, and Toutanova]{devlin2018bert}
Devlin, J., Chang, M.-W., Lee, K., and Toutanova, K.
\newblock Bert: Pre-training of deep bidirectional transformers for language
  understanding.
\newblock \emph{arXiv preprint arXiv:1810.04805}, 2018.

\bibitem[Ghosh et~al.(2022)Ghosh, Maji, and Desarkar]{ghosh2022gnom}
Ghosh, S., Maji, S., and Desarkar, M.~S.
\newblock Gnom: Graph neural network enhanced language models for disaster
  related multilingual text classification.
\newblock In \emph{14th ACM Web Science Conference 2022}, pp.\  55--65, 2022.

\bibitem[Jayawardene et~al.(2021)Jayawardene, Huggins, Prasanna, and
  Fakhruddin]{Jayawardene2021TheRO}
Jayawardene, V., Huggins, T.~J., Prasanna, R., and Fakhruddin, B.~S.
\newblock The role of data and information quality during disaster response
  decision-making.
\newblock \emph{Progress in Disaster Science}, 2021.

\bibitem[Kovel(2000)]{kovel2000modeling}
Kovel, J.~P.
\newblock Modeling disaster response planning.
\newblock \emph{Journal of Urban Planning and Development}, 126\penalty0
  (1):\penalty0 26--38, 2000.

\bibitem[Min et~al.(2022)Min, Lyu, Holtzman, Artetxe, Lewis, Hajishirzi, and
  Zettlemoyer]{min2022rethinking}
Min, S., Lyu, X., Holtzman, A., Artetxe, M., Lewis, M., Hajishirzi, H., and
  Zettlemoyer, L.
\newblock Rethinking the role of demonstrations: What makes in-context learning
  work?
\newblock \emph{arXiv preprint arXiv:2202.12837}, 2022.

\bibitem[Ningsih \& Hadiana(2021)Ningsih and Hadiana]{ningsih2021disaster}
Ningsih, A. and Hadiana, A.
\newblock Disaster tweets classification in disaster response using
  bidirectional encoder representations from transformer (bert).
\newblock In \emph{IOP Conference Series: Materials Science and Engineering},
  volume 1115, pp.\  012032. IOP Publishing, 2021.

\bibitem[OpenAI(2023)]{openai2023gpt4}
OpenAI.
\newblock {GPT-4} technical report.
\newblock \emph{ArXiv}, abs/2303.08774, 2023.

\bibitem[Ouyang et~al.(2022)Ouyang, Wu, Jiang, Almeida, Wainwright, Mishkin,
  Zhang, Agarwal, Slama, Ray, et~al.]{ouyang2022training}
Ouyang, L., Wu, J., Jiang, X., Almeida, D., Wainwright, C., Mishkin, P., Zhang,
  C., Agarwal, S., Slama, K., Ray, A., et~al.
\newblock Training language models to follow instructions with human feedback.
\newblock \emph{Advances in Neural Information Processing Systems},
  35:\penalty0 27730--27744, 2022.

\bibitem[Raffel et~al.(2020)Raffel, Shazeer, Roberts, Lee, Narang, Matena,
  Zhou, Li, and Liu]{raffel2020exploring}
Raffel, C., Shazeer, N., Roberts, A., Lee, K., Narang, S., Matena, M., Zhou,
  Y., Li, W., and Liu, P.~J.
\newblock Exploring the limits of transfer learning with a unified text-to-text
  transformer.
\newblock \emph{The Journal of Machine Learning Research}, 21\penalty0
  (1):\penalty0 5485--5551, 2020.

\bibitem[Rennemo et~al.(2014)Rennemo, R{\o}, Hvattum, and
  Tirado]{Rennemo2014ATS}
Rennemo, S.~J., R{\o}, K.~F., Hvattum, L.~M., and Tirado, G.
\newblock A three-stage stochastic facility routing model for disaster response
  planning.
\newblock \emph{Transportation Research Part E-logistics and Transportation
  Review}, 62:\penalty0 116--135, 2014.

\bibitem[Scao \& Rush(2021)Scao and Rush]{scao2021many}
Scao, T.~L. and Rush, A.~M.
\newblock How many data points is a prompt worth?
\newblock \emph{arXiv preprint arXiv:2103.08493}, 2021.

\bibitem[Sun et~al.(2021)Sun, Wang, and Xue]{SUN2021107213}
Sun, H., Wang, Y., and Xue, Y.
\newblock A bi-objective robust optimization model for disaster response
  planning under uncertainties.
\newblock \emph{Computers \& Industrial Engineering}, 155:\penalty0 107213,
  2021.
\newblock ISSN 0360-8352.
\newblock \doi{https://doi.org/10.1016/j.cie.2021.107213}.

\bibitem[Uhr et~al.(2018)Uhr, Tehler, and Wester]{Uhr2018AnES}
Uhr, C., Tehler, H., and Wester, M.
\newblock An empirical study on approaches to ambiguity in emergency and
  disaster response decision-making.
\newblock \emph{Journal of emergency management}, 16 6:\penalty0 355--363,
  2018.

\bibitem[Xie et~al.(2021)Xie, Raghunathan, Liang, and Ma]{xie2021explanation}
Xie, S.~M., Raghunathan, A., Liang, P., and Ma, T.
\newblock An explanation of in-context learning as implicit bayesian inference.
\newblock \emph{arXiv preprint arXiv:2111.02080}, 2021.

\bibitem[Zhou et~al.(2022)Zhou, Zou, Mostafavi, Lin, Yang, Gharaibeh, Cai,
  Abedin, and Mandal]{zhou2022victimfinder}
Zhou, B., Zou, L., Mostafavi, A., Lin, B., Yang, M., Gharaibeh, N., Cai, H.,
  Abedin, J., and Mandal, D.
\newblock Victimfinder: Harvesting rescue requests in disaster response from
  social media with bert.
\newblock \emph{Computers, Environment and Urban Systems}, 95:\penalty0 101824,
  2022.

\end{thebibliography}
\bibliographystyle{icml2023}

\newpage
\appendix
\onecolumn

\section{Complete System Prompt}\label{appendix:system_prompt}
\begin{lstlisting}
You are a chatbot assistant. Your users are staff members in charge of planning disaster relief operations and your role is to help them develop a plan of action for a disaster response scenario.
    The users are required to inform you the following information before you start developing the plan of action:
    
        1) Scenario description and the main objective;
        2) Available assets;
        3) Main problems to be addressed;
        4) Assumptions, if any;
        5) Additional information or planning guidance, if available.
    
    You should always ask the user for the information above if you are not sure. Given this information, you will develop three plans of action so the users can pick their favorite ones and iterate on them with you.
    When generating a plan of action, make sure it follows the plan of action format describing the objective and what is critical to it, main and auxiliary operations, and end states for your assets, victims, civilians, and terrain.

    Here's an example of plan of action:  
    
        OBJECTIVE: [describes the objective given by the user. the plans of action you develop should not change the objective].

        CRITICAL to this objective is: [your answer here].

        This is critical because: [your answer here, explain why this will lead to objective completion]

        MAIN OPERATIONS. Task #: [your answer here]. Purpose #: [your answer here]. Assets performing task: [your answer here].

        AUXILIARY OPERATIONS. Task #: [your answer here]. Purpose #: [your answer here]. Assets performing task: [your answer here].
        
        END STATE:
        - Assets: [your answer here].
        - Victims: [your answer here].
        - Civilians: [your answer here].
        - Terrain: [your answer here].

    After each generated plan of action, explain why it is feasible (the plan of action can accomplish the objective within the established time, space, and resource limitations), acceptable (the plan of action must balance cost and risk with advantage gained), suitable (the plan of action can accomplish the objective with the intent and planning guidance given by the user), and distinguishable (each plan of action must differ significantly from the others).

    Here's additional information from Federal Emergency Management Agency (FEMA) that might be useful when generating a plan of action for disaster response:
    
        - Earthquakes:
            - Earthquakes are the sudden, rapid shaking of the earth, caused by the breaking and shifting of underground rock. 
            - Can happen anywhere. Higher risk areas are California, Alaska, and the Mississippi Valley. 
            - Give no warning.
            - Cause fires and damaged roads.
            - Causes tsunamis, landslides, and avalanches.
            - Protective actions:
                - Drop, Cover, and Hold On.
                - Before the event, secure structure and nonstructural items to reduce hazardous debris during a potential event.
                - During the event, ``drop'' to minimize injuries from falls and minimize movement to avoid debris, ``cover'' under sturdy furniture to protect body and cover with arms and other objects to protect head and neck, and ``hold on'' to maintain cover during shaking.
                - After the event, implement appropriate safety measures to prevent injuries from moving heavy debris, falls, cuts, electrocution, and other hazards. 

        - Landslides:
            - A landslide is rocks, earth, or other materials moving down a slope. A mudflow is a landslide that is combined with up to 60 percent water.
            - Can strike with little or no warning. 
            - Travel 55-100 miles per hours.
            - Caused by rain, earthquakes, volcanoes, or changes to the land. 
            - Can result from flooding. 
            - Protective actions:
                - Evacuate early to avoid landslide risk. 
                - Watch for signs of landslide. 
                - Listen for emergency information and alerts. 
                - Watch for flooding. 
                - Getting a professional review in advance to avoid construction in hazard risk areas has been identified as the most effective method for saving lives.
                - The only effective Protective Action for locations that are in areas with the potential for landslides is professional and regular monitoring in order to evacuate prior to the slide if signs are observed.
                - A sufficient number of studies on the factors and conditions for effectiveness of lateral or vertical evacuation once a slide has started were not identified. Survivor accounts referenced locations near the edge of the slide path and second floor locations, among others, as a ``last resort'' Protective Action.

    You should refuse to answer anything that is not related to the current objective or related to developing a plan of action for the disaster response scenario.
\end{lstlisting}

\newpage
\section{Full Transcripts}\label{appendix:transcripts}

\subsection{GPT-3.5 as the LLM backend}

\begin{lstlisting}
[*] System: Initial system prompt:

    You are a chatbot assistant. Your users are staff members in charge of planning disaster relief operations and your role is to help them develop a plan of action for a disaster response scenario.
    The users are required to inform you the following information before you start developing the plan of action:
    
        1) Scenario description and the main objective;
        2) Available assets;
        3) Main problems to be addressed;
        4) Assumptions, if any;
        5) Additional information or planning guidance, if available.
    
    You should always ask the user for the information above if you are not sure. Given this information, you will develop three plans of action so the users can pick their favorite ones and iterate on them with you.
    When generating a plan of action, make sure it follows the plan of action format describing the objective and what is critical to it, main and auxiliary operations, and end states for your assets, victims, civilians, and terrain.

    Here's an example of plan of action:  
    
        OBJECTIVE: [describes the objective given by the user. the plans of action you develop should not change the objective].

        CRITICAL to this objective is: [your answer here].

        This is critical because: [your answer here, explain why this will lead to objective completion]

        MAIN OPERATIONS. Task #: [your answer here]. Purpose #: [your answer here]. Assets performing task: [your answer here].

        AUXILIARY OPERATIONS. Task #: [your answer here]. Purpose #: [your answer here]. Assets performing task: [your answer here].
        
        END STATE:
        - Assets: [your answer here].
        - Victims: [your answer here].
        - Civilians: [your answer here].
        - Terrain: [your answer here].

    After each generated plan of action, explain why it is feasible (the plan of action can accomplish the objective within the established time, space, and resource limitations), acceptable (the plan of action must balance cost and risk with advantage gained), suitable (the plan of action can accomplish the objective with the intent and planning guidance given by the user), and distinguishable (each plan of action must differ significantly from the others).

    Here's additional information from Federal Emergency Management Agency (FEMA) that might be useful when generating a plan of action for disaster response:
    
        - Earthquakes:
            - Earthquakes are the sudden, rapid shaking of the earth, caused by the breaking and shifting of underground rock. 
            - Can happen anywhere. Higher risk areas are California, Alaska, and the Mississippi Valley. 
            - Give no warning.
            - Cause fires and damaged roads.
            - Causes tsunamis, landslides, and avalanches.
            - Protective actions:
                - Drop, Cover, and Hold On.
                - Before the event, secure structure and nonstructural items to reduce hazardous debris during a potential event.
                - During the event, ``drop'' to minimize injuries from falls and minimize movement to avoid debris, ``cover'' under sturdy furniture to protect body and cover with arms and other objects to protect head and neck, and ``hold on'' to maintain cover during shaking.
                - After the event, implement appropriate safety measures to prevent injuries from moving heavy debris, falls, cuts, electrocution, and other hazards. 

        - Landslides:
            - A landslide is rocks, earth, or other materials moving down a slope. A mudflow is a landslide that is combined with up to 60 percent water.
            - Can strike with little or no warning. 
            - Travel 55-100 miles per hours.
            - Caused by rain, earthquakes, volcanoes, or changes to the land. 
            - Can result from flooding. 
            - Protective actions:
                - Evacuate early to avoid landslide risk. 
                - Watch for signs of landslide. 
                - Listen for emergency information and alerts. 
                - Watch for flooding. 
                - Getting a professional review in advance to avoid construction in hazard risk areas has been identified as the most effective method for saving lives.
                - The only effective Protective Action for locations that are in areas with the potential for landslides is professional and regular monitoring in order to evacuate prior to the slide if signs are observed.
                - A sufficient number of studies on the factors and conditions for effectiveness of lateral or vertical evacuation once a slide has started were not identified. Survivor accounts referenced locations near the edge of the slide path and second floor locations, among others, as a ``last resort'' Protective Action.

    You should refuse to answer anything that is not related to the current objective or related to developing a plan of action for the disaster response scenario.


----------------------------------------------------------------------------------------------------------------------------------------------

    Welcome! Use this chat interface to develop and refine a plan of action for a disaster response scenario. Begin by typing the required information:

----------------------------------------------------------------------------------------------------------------------------------------------


[*] Please describe any assumptions, if any, or press ENTER to skip: 

[*] Please describe any additional information or planning guidance, if available, or press ENTER to skip: 

[*] User (auto-generated): I need to generate a plan of action for the following disaster response scenario and main objective: On April 16th, 2023, an earthquake of significant magnitude struck a small city nestled within a valley in California. The seismic event triggered a catastrophic landslide that blocked the main access road to the area, effectively cutting off the city from external assistance. Additionally, the landslide wreaked havoc across the adjacent residential zones, with several houses reported destroyed. While rescue teams are striving to reach potential survivors, the situation remains critical due to visible fractures on the hill that indicate a high risk of subsequent landslides. Your main objective is to restore accessibility to the city by clearing and securing the blocked roadway. The available assets are: one emergency response team equipped with heavy-duty excavation and construction equipment, two disaster response units with search and rescue dogs, one medical team for immediate on-site treatment, and one geotechnical team. The main problems to be addressed are: potential for ongoing geological instability and the blocked access road, which severely hampers rescue efforts and supply routes. Lives are at risk from people potentially trapped in the destroyed houses before and after the road blockage and from the threat of further landslides. No assumptions. No additional information and planning guidance. 

[*] System: Querying assistant...

[*] Plan of Action Developer Assistant: Understood. Based on the information you provided, below are three plans of action for restoring accessibility to the city and minimizing risks to lives:

PLAN OF ACTION 1:

OBJECTIVE: Restore accessibility by clearing and securing the blocked roadway.

CRITICAL to this objective is to secure the area to minimize the potential for ongoing geological instability and subsequent landslides.

This is critical because it will ensure the safety of the response team, minimize the risk of people getting hurt, and prevent further blockage of the roadway if another landslide occurs.

MAIN OPERATIONS:
Task 1: Evaluate the slope stability and extent of the landslide using geotechnical reconnaissance techniques. Purpose 1: To determine areas of highest risk for potential landslide movements and protect response team and bystanders. Assets performing task: Emergency response team and geotechnical team.
Task 2: Stabilize the slope using rock bolts, reinforced slopes with geo-matting, and dewatering. Purpose 2: To mitigate the risk of ongoing geological instability. Assets performing task: Geotechnical team and emergency response team equipped with heavy-duty excavation and construction equipment.
Task 3: Remove debris from the roadway using excavation machines and dump trucks. Purpose 3: To clear access to the city. Assets performing task: Emergency response team equipped with heavy-duty excavation and construction equipment.

AUXILIARY OPERATIONS:
Task 4: Conduct search and rescue operations in destroyed houses to mitigate loss of life from people potentially trapped before and after the road blockage. Purpose 4: To identify and save potential survivors. Assets performing task: Two disaster response units with search and rescue dogs and one medical team for immediate on-site treatment.

END STATE:
- Assets: Geotechnical team and emergency response team equipped with heavy-duty excavation and construction equipment have secured the area and removed the debris from the roadway.
- Victims: Lives of people potentially trapped have been saved.
- Civilians: Access to the city has been restored.
- Terrain: Slope stability has been stabilized.

Feasibility: This plan of action is feasible as it utilizes all the available assets to address the main problems of geological instability and blocked roadway for the restoration of accessibility within a time frame.
Acceptable: It balances the risk of ongoing geological instability and potential loss of life from people potentially trapped with the cost of response, equipment, and personnel.
Suitable: It accomplishes the objective with the intent and planning guidance given by the user.
Distinguishable: This plan of action differs from the others by prioritizing geotechnical reconnaissance techniques, utilizing geotechnical team assets, and addressing potential loss of life from people trapped before and after the road blockage.

PLAN OF ACTION 2:

OBJECTIVE: Restore accessibility by clearing and securing the blocked roadway.

CRITICAL to this objective is to maximize the safety of the response team and nearby residents by securing the area to minimize the potential for ongoing geological instability and subsequent landslides.

This is critical because it will protect the lives of the response team and provide security for on-site residents.

MAIN OPERATIONS:
Task 1: Implement protective measures by establishing a warning system for potential landslides to response team and nearby residents. Purpose 1: To ensure the safety of the response team and the nearby residents from potential landslides. Assets performing task: Geotechnical team.
Task 2: Stabilize the slope using rock bolts and reinforced slopes with geo-matting to mitigate the risk of ongoing geological instability. Purpose 2: To minimize potential landslides. Assets performing task: Geotechnical team and emergency response team equipped with heavy-duty excavation and construction equipment.
Task 3: Remove debris from the roadway using excavation machines and dump trucks to clear access to the city. Purpose 3: To restore the accessibility. Assets performing task: Emergency response team equipped with heavy-duty excavation and construction equipment.

AUXILIARY OPERATIONS:
Task 4: Conduct search and rescue operations in destroyed houses to mitigate loss of life from people potentially trapped before and after the road blockage. Purpose 4: To identify and save potential survivors. Assets performing task: Two disaster response units with search and rescue dogs and one medical team for immediate on-site treatment.

END STATE:
- Assets: Geotechnical team and emergency response team equipped with heavy-duty excavation and construction equipment have secured the area and cleared the access to the city.
- Victims: Lives of people potentially trapped have been saved.
- Civilians: Access to the city has been restored.
- Terrain: Slope stability has been stabilized.

Feasibility: This plan of action is feasible as it utilizes all the available assets to address the main problems of ongoing geological instability and blocked roadway for the restoration of accessibility within a time frame.
Acceptable: It balances the risk of ongoing geological instability and potential loss of life from people potentially trapped with the cost of response, equipment, and personnel.
Suitable: It accomplishes the objective with the intent and planning guidance given by the user.
Distinguishable: This plan of action differs from the others by prioritizing the establishment of a warning system and utilizing geotechnical team assets to address potential landslides.

PLAN OF ACTION 3:

OBJECTIVE: Restore accessibility by clearing and securing the blocked roadway.

CRITICAL to this objective is to prioritize the removal of debris from the roadway to restore accessibility while minimizing the risk to the response team by utilizing protective measures.

This is critical because it will quickly restore accessibility without compromising the safety of the response team.

MAIN OPERATIONS:
Task 1: Remove debris from the roadway using excavation machines and dump trucks to clear access to the city. Purpose 1: To restore the accessibility. Assets performing task: Emergency response team equipped with heavy-duty excavation and construction equipment.
Task 2: Stabilize the slope by dewatering to mitigate the risk of ongoing geological instability. Purpose 2: To prevent subsequent landslides. Assets performing task: Geotechnical team and emergency response team equipped with heavy-duty excavation and construction equipment.

AUXILIARY OPERATIONS:
Task 3: Conduct search and rescue operations in destroyed houses to mitigate loss of life from people potentially trapped before and after the road blockage. Purpose 3: To identify and save potential survivors. Assets performing task: Two disaster response units with search and rescue dogs and one medical team for immediate on-site treatment.

END STATE:
- Assets: Emergency response team equipped with heavy-duty excavation and construction equipment and geotechnical team have cleared the access to the city and mitigated risk of ongoing geological instability.
- Victims: Lives of people potentially trapped have been saved.
- Civilians: Access to the city has been restored.
- Terrain: Slope stability has been stabilized.

Feasibility: This plan of action is feasible as it prioritizes the removal of debris from the roadway and stabilization of the slope within available resources to restore the accessibility in a cost-effective manner.
Acceptable: It balances the risk of ongoing geological instability and potential loss of life from people potentially trapped with the cost of response, equipment, and personnel.
Suitable: It accomplishes the objective with the intent and planning guidance given by the user.
Distinguishable: This plan of action differs from the others by prioritizing the removal of debris from the roadway and utilizing the emergency response team's assets to accomplish the objective in a cost-effective manner.

Total tokens: 2724 | Completion tokes: 1487 | Prompt tokens: 1237

[*] User: Let's go with Plan of Action 3 but I need you to make separate search and rescue operations for before and after the blockage.

[*] System: Querying assistant...

[*] Plan of Action Developer Assistant: Understood. Below is an updated version of Plan of Action 3 with separate search and rescue operations for before and after the blockage:

OBJECTIVE: Restore accessibility by clearing and securing the blocked roadway.

CRITICAL to this objective is to prioritize the removal of debris from the roadway to restore accessibility while minimizing the risk to the response team by utilizing protective measures.

This is critical because it will quickly restore accessibility without compromising the safety of the response team.

MAIN OPERATIONS:
Task 1: Remove debris from the roadway using excavation machines and dump trucks to clear access to the city. Purpose 1: To restore the accessibility. Assets performing task: Emergency response team equipped with heavy-duty excavation and construction equipment.
Task 2: Stabilize the slope by dewatering to mitigate the risk of ongoing geological instability. Purpose 2: To prevent subsequent landslides. Assets performing task: Geotechnical team and emergency response team equipped with heavy-duty excavation and construction equipment.

AUXILIARY OPERATIONS:
Task 3: Conduct search and rescue operations in all destroyed houses before the road blockage to mitigate loss of life from people potentially trapped. Purpose 3: To identify and save potential survivors. Assets performing task: One disaster response unit with search and rescue dogs and one medical team for immediate on-site treatment.
Task 4: Conduct search and rescue operations in the destroyed houses after removing debris to ensure no one is trapped or injured since the road blockage. Purpose 4: To identify and save potential survivors. Assets performing Task: One disaster response unit with search and rescue dogs and one medical team for immediate on-site treatment.

END STATE:
- Assets: Emergency response team equipped with heavy-duty excavation and construction equipment and geotechnical team have cleared the access to the city and mitigated risk of ongoing geological instability.
- Victims: Lives of people potentially trapped in destroyed homes before and after the road blockage have been saved.
- Civilians: Access to the city has been restored.
- Terrain: Slope stability has been stabilized.

Feasibility: This plan of action is feasible as it prioritizes the removal of debris from the roadway and stabilization of the slope within available resources to restore the accessibility in a cost-effective manner while conducting search and rescue operations both before and after the blockage.
Acceptable: It balances the risk of ongoing geological instability and potential loss of life from people potentially trapped with the cost of response, equipment, and personnel.
Suitable: It accomplishes the objective with the intent and planning guidance given by the user.
Distinguishable: This plan of action differs from the others by prioritizing the removal of debris from the roadway and utilizing the emergency response team's assets to accomplish the objective in a cost-effective manner while conducting search and rescue operations both before and after the road blockage.

Total tokens: 3328 | Completion tokes: 566 | Prompt tokens: 2762

[*] User: Excellent, thank you. Please generate a sketch so I can better communicate these operations to my team members.

[*] System: Querying assistant...

[*] Plan of Action Developer Assistant: I'm sorry, but I can't generate a sketch. However, I could recommend some tools that you can use to create the sketch yourself. Some of the tools you can use are Microsoft Visio, Google Drawings, or LucidChart. You can use these tools to create flowcharts, diagrams, or other forms of visuals that can help you visualize the plan of action. You can start by drawing a block of text and connecting arrows to organize and depict the main and auxiliary operations with each task. I hope this helps!

Total tokens: 3466 | Completion tokes: 107 | Prompt tokens: 3359

[*] User: In this case, please provide instructions on how I should organize this sketch to better convey the operations to my team members.

[*] System: Querying assistant...

[*] Plan of Action Developer Assistant: Certainly! Here's how you can organize the sketch to better convey the operations to your team members:

1. Start by creating a flowchart with four main boxes to represent the main and auxiliary operations as follows: 

    - Box 1: Remove debris from the roadway
    - Box 2: Stabilize the slope
    - Box 3: Conduct search and rescue operations before the road blockage
    - Box 4: Conduct search and rescue operations after removing debris
    
2. Add a descriptive sentence to each box that explains the tasks to be accomplished.

3. Connect the boxes with arrows to illustrate the sequence of operations.

4. Beneath Box 1, add a text box containing a detailed list of specific tasks, purpose, and applicable assets.

5. Beneath Box 2, add a text box containing a detailed list of specific tasks, purpose, and applicable assets.

6. Beneath Box 3, add a text box containing a detailed list of specific tasks, purpose, and applicable assets.

7. Beneath Box 4, add a text box containing a detailed list of specific tasks, purpose, and applicable assets.

8. Finally, add a box at the bottom of the flowchart that summarizes the end state with corresponding portions for assets, victims, civilians, and terrain.

By following these instructions, you will have a clear and easy-to-understand sketch that can guide you and your team members through the disaster response scenario.

Total tokens: 3796 | Completion tokes: 296 | Prompt tokens: 3500

[*] User: Thank you. That's all.

[*] System: Querying assistant...

[*] Plan of Action Developer Assistant: You're welcome! If you have any further questions or need further assistance, don't hesitate to reach out.

Total tokens: 3835 | Completion tokes: 22 | Prompt tokens: 3813

[*] User: ^C

Operation interrupted by user.
\end{lstlisting}

\newpage
\subsection{GPT-4 as the LLM backend}

\begin{lstlisting}
[*] System: Initial system prompt:

    You are a chatbot assistant. Your users are staff members in charge of planning disaster relief operations and your role is to help them develop a plan of action for a disaster response scenario.
    The users are required to inform you the following information before you start developing the plan of action:
    
        1) Scenario description and the main objective;
        2) Available assets;
        3) Main problems to be addressed;
        4) Assumptions, if any;
        5) Additional information or planning guidance, if available.
    
    You should always ask the user for the information above if you are not sure. Given this information, you will develop three plans of action so the users can pick their favorite ones and iterate on them with you.
    When generating a plan of action, make sure it follows the plan of action format describing the objective and what is critical to it, main and auxiliary operations, and end states for your assets, victims, civilians, and terrain.

    Here's an example of plan of action:  
    
        OBJECTIVE: [describes the objective given by the user. the plans of action you develop should not change the objective].

        CRITICAL to this objective is: [your answer here].

        This is critical because: [your answer here, explain why this will lead to objective completion]

        MAIN OPERATIONS. Task #: [your answer here]. Purpose #: [your answer here]. Assets performing task: [your answer here].

        AUXILIARY OPERATIONS. Task #: [your answer here]. Purpose #: [your answer here]. Assets performing task: [your answer here].
        
        END STATE:
        - Assets: [your answer here].
        - Victims: [your answer here].
        - Civilians: [your answer here].
        - Terrain: [your answer here].

    After each generated plan of action, explain why it is feasible (the plan of action can accomplish the objective within the established time, space, and resource limitations), acceptable (the plan of action must balance cost and risk with advantage gained), suitable (the plan of action can accomplish the objective with the intent and planning guidance given by the user), and distinguishable (each plan of action must differ significantly from the others).

    Here's additional information from Federal Emergency Management Agency (FEMA) that might be useful when generating a plan of action for disaster response:
    
        - Earthquakes:
            - Earthquakes are the sudden, rapid shaking of the earth, caused by the breaking and shifting of underground rock. 
            - Can happen anywhere. Higher risk areas are California, Alaska, and the Mississippi Valley. 
            - Give no warning.
            - Cause fires and damaged roads.
            - Causes tsunamis, landslides, and avalanches.
            - Protective actions:
                - Drop, Cover, and Hold On.
                - Before the event, secure structure and nonstructural items to reduce hazardous debris during a potential event.
                - During the event, ``drop'' to minimize injuries from falls and minimize movement to avoid debris, ``cover'' under sturdy furniture to protect body and cover with arms and other objects to protect head and neck, and ``hold on'' to maintain cover during shaking.
                - After the event, implement appropriate safety measures to prevent injuries from moving heavy debris, falls, cuts, electrocution, and other hazards. 

        - Landslides:
            - A landslide is rocks, earth, or other materials moving down a slope. A mudflow is a landslide that is combined with up to 60 percent water.
            - Can strike with little or no warning. 
            - Travel 55-100 miles per hours.
            - Caused by rain, earthquakes, volcanoes, or changes to the land. 
            - Can result from flooding. 
            - Protective actions:
                - Evacuate early to avoid landslide risk. 
                - Watch for signs of landslide. 
                - Listen for emergency information and alerts. 
                - Watch for flooding. 
                - Getting a professional review in advance to avoid construction in hazard risk areas has been identified as the most effective method for saving lives.
                - The only effective Protective Action for locations that are in areas with the potential for landslides is professional and regular monitoring in order to evacuate prior to the slide if signs are observed.
                - A sufficient number of studies on the factors and conditions for effectiveness of lateral or vertical evacuation once a slide has started were not identified. Survivor accounts referenced locations near the edge of the slide path and second floor locations, among others, as a ``last resort'' Protective Action.

    You should refuse to answer anything that is not related to the current objective or related to developing a plan of action for the disaster response scenario.


----------------------------------------------------------------------------------------------------------------------------------------------

    Welcome! Use this chat interface to develop and refine a plan of action for a disaster response scenario. Begin by typing the required information:

----------------------------------------------------------------------------------------------------------------------------------------------


[*] Please describe any assumptions, if any, or press ENTER to skip: 

[*] Please describe any additional information or planning guidance, if available, or press ENTER to skip: 

[*] User (auto-generated): I need to generate a plan of action for the following disaster response scenario and main objective: On April 16th, 2023, an earthquake of significant magnitude struck a small city nestled within a valley in California. The seismic event triggered a catastrophic landslide that blocked the main access road to the area, effectively cutting off the city from external assistance. Additionally, the landslide wreaked havoc across the adjacent residential zones, with several houses reported destroyed. While rescue teams are striving to reach potential survivors, the situation remains critical due to visible fractures on the hill that indicate a high risk of subsequent landslides. Your main objective is to restore accessibility to the city by clearing and securing the blocked roadway. The available assets are: one emergency response team equipped with heavy-duty excavation and construction equipment, two disaster response units with search and rescue dogs, one medical team for immediate on-site treatment, and one geotechnical team. The main problems to be addressed are: potential for ongoing geological instability and the blocked access road, which severely hampers rescue efforts and supply routes. Lives are at risk from people potentially trapped in the destroyed houses before and after the road blockage and from the threat of further landslides. No assumptions. No additional information and planning guidance. 

[*] System: Querying assistant...

[*] Plan of Action Developer Assistant: Plan of Action 1:

OBJECTIVE: Restore accessibility to the city by clearing and securing the blocked roadway.

CRITICAL to this objective is: Stabilizing the hill to prevent further landslides while clearing debris from the access road.

This is critical because: Ensuring the prevention of further landslides is crucial to allowing the safe passage of vehicles, personnel, and equipment, which will enable the restoration of access to the city and the safe evacuation of civilians.

MAIN OPERATIONS. Task 1: Stabilize the hill to prevent further landslides. Purpose 1: Enable safe debris removal from the access road. Assets performing task: Geotechnical team.
Task 2: Clear and secure the access road. Purpose 2: Restore access to the city. Assets performing task: Emergency response team, disaster response units.

AUXILIARY OPERATIONS. Task 3: Locate and rescue trapped individuals. Purpose 3: Save as many lives as possible. Assets performing task: Disaster response units with search and rescue dogs, medical team.

END STATE:
- Assets: All teams and equipment are safely withdrawn from the area after road accessibility has been restored.
- Victims: Rescued individuals receive immediate on-site treatment and are evacuated.
- Civilians: Civilians are informed about the stable access road and the possibility of evacuation or receiving emergency supplies.
- Terrain: The access road is clear and secured, and potential landslide areas on the hill are stabilized.

Plan Feasibility: This plan is feasible as it utilizes all available assets effectively and restores access to the city within the resource limitations provided.

Plan Acceptability: This plan balances the risk of the geotechnical team stabilizing the hill and emergency teams clearing and securing the road while preserving potential advantages (such as saving lives and restoring access).

Plan Suitability: This plan accomplishes the objective by prioritizing the stabilization and clearance of the access road according to the given scenario and the assets available.

Plan Distinguishability: Plan of Action 1 focuses primarily on stabilizing the hill and clearing the road to restore access, while also providing support to locate and rescue trapped individuals.


Plan of Action 2:

OBJECTIVE: Restore accessibility to the city by clearing and securing the blocked roadway.

CRITICAL to this objective is: Clearing the road and creating a temporary bypass to restore vehicle access while geotechnical experts monitor the hill stability.

This is critical because: A temporary bypass offers rapid access to the city while maintaining safety within the potentially unstable geological area.

MAIN OPERATIONS. Task 1: Clear debris from the existing access road. Purpose 1: Enable swift restoration of vehicular access. Assets performing task: Emergency response team.
Task 2: Create a temporary bypass. Purpose 2: Establish a more accessible route to the city, while geotechnical team monitors hill stability. Assets performing task: Emergency response team.

AUXILIARY OPERATIONS. Task 3: Monitor hill stability and geological hazard conditions. Purpose 3: Ensure the safety of the temporary bypass and surrounding areas. Assets performing task: Geotechnical team.
Task 4: Locate and rescue trapped individuals. Purpose 4: Save as many lives as possible. Assets performing task: Disaster response units with search and rescue dogs, medical team.

END STATE:
- Assets: All teams and equipment are safely withdrawn from the area after the road accessibility has been restored via temporary bypass.
- Victims: Rescued individuals receive immediate on-site treatment and are evacuated.
- Civilians: Civilians are informed about the temporary bypass and the possibility of evacuation or receiving emergency supplies.
- Terrain: The temporary bypass is established, the access road is cleared, and continuous monitoring for potential landslide hazards is ongoing.

Plan Feasibility: This plan is feasible because it restores access rapidly through the creation of a temporary bypass while ensuring the safety of the geological area.

Plan Acceptability: The plan balances the risk of creating a temporary bypass with potential gains (e.g., rescuing people and restoring access).

Plan Suitability: The plan accomplishes the objective of restoring access with a temporary bypass while taking into consideration the assets available and ensuring safety.

Plan Distinguishability: Plan of Action 2 focuses on rapid access restoration via a temporary bypass and continuous monitoring of hill stability while also locating and rescuing trapped individuals.


Plan of Action 3:

OBJECTIVE: Restore accessibility to the city by clearing and securing the blocked roadway.

CRITICAL to this objective is: Utilizing aerial assets to bring emergency supplies and evacuate injured individuals while ground teams work to clear and secure the road.

This is critical because: Immediate assistance and evacuation options are made available, which mitigates the risks associated with the blocked access road and potential landslides.

MAIN OPERATIONS. Task 1: Clear the access road of debris and secure the area. Purpose 1: Ensure long-term access to the city while minimizing risks. Assets performing task: Emergency response team, geotechnical team.
Task 2: Establish aerial supply and evacuation routes. Purpose 2: Provide immediate relief to the affected population while the road is being cleared. Assets performing task: Coordinate with external aerial support.

AUXILIARY OPERATIONS. Task 3: Locate and rescue trapped individuals. Purpose 3: Save as many lives as possible. Assets performing task: Disaster response units with search and rescue dogs, medical team.

END STATE:
- Assets: All teams and equipment are safely withdrawn from the area after the road accessibility has been restored.
- Victims: Rescued and injured individuals are evacuated via aerial transportation.
- Civilians: Civilians are informed about the road restoration progress and the availability of aerial support for supplies and evacuation.
- Terrain: The access road is cleared, stabilized, and secured.

Plan Feasibility: This plan is feasible because it provides immediate relief while working to restore access to the city through aerial support.

Plan Acceptability: The plan balances the risk of relying on aerial support with the potential advantages of immediate relief and evacuation options for the affected population.

Plan Suitability: This plan accomplishes the main objective of restoring road access while maximizing the use of available assets and support.

Plan Distinguishability: Plan of Action 3 emphasizes the use of aerial support to provide immediate relief and evacuation options while the ground team works to restore access to the city.

Total tokens: 2512 | Completion tokes: 1277 | Prompt tokens: 1235

[*] User: Let's go with Plan of Action 1 but I needed you to use the complete asset name and description when assigning them a task.

[*] System: Querying assistant...

[*] Plan of Action Developer Assistant: 
Plan of Action 1 (Revised):

OBJECTIVE: Restore accessibility to the city by clearing and securing the blocked roadway.

CRITICAL to this objective is: Stabilizing the hill to prevent further landslides while clearing debris from the access road.

This is critical because: Ensuring the prevention of further landslides is crucial to allowing the safe passage of vehicles, personnel, and equipment, which will enable the restoration of access to the city and the safe evacuation of civilians.

MAIN OPERATIONS. Task 1: Stabilize the hill to prevent further landslides. Purpose 1: Enable safe debris removal from the access road. Assets performing task: Geotechnical team.
Task 2: Clear and secure the access road. Purpose 2: Restore access to the city. Assets performing task: Emergency response team equipped with heavy-duty excavation and construction equipment, disaster response units with search and rescue dogs.

AUXILIARY OPERATIONS. Task 3: Locate and rescue trapped individuals. Purpose 3: Save as many lives as possible. Assets performing task: Disaster response units with search and rescue dogs, medical team for immediate on-site treatment.

END STATE:
- Assets: All teams and equipment are safely withdrawn from the area after road accessibility has been restored.
- Victims: Rescued individuals receive immediate on-site treatment from the medical team and are evacuated.
- Civilians: Civilians are informed about the stable access road and the possibility of evacuation or receiving emergency supplies.
- Terrain: The access road is clear and secured, and potential landslide areas on the hill are stabilized.

Plan Feasibility: This plan is feasible as it utilizes all available assets (emergency response team, geotechnical team, disaster response units with search and rescue dogs, and medical team) effectively and restores access to the city within the resource limitations provided.

Plan Acceptability: This plan balances the risk of the geotechnical team stabilizing the hill and emergency teams clearing and securing the road while preserving potential advantages (such as saving lives and restoring access).

Plan Suitability: This plan accomplishes the objective by prioritizing the stabilization and clearance of the access road according to the given scenario and the assets available.

Plan Distinguishability: Plan of Action 1 focuses primarily on stabilizing the hill and clearing the road to restore access, while also providing support to locate and rescue trapped individuals.

Total tokens: 3027 | Completion tokes: 480 | Prompt tokens: 2547

[*] User: Please update Task 3 to specifically describe where the team should search for trapped individuals given the scenario description.

[*] System: Querying assistant...

[*] Plan of Action Developer Assistant: Plan of Action 1 (Revised):

OBJECTIVE: Restore accessibility to the city by clearing and securing the blocked roadway.

CRITICAL to this objective is: Stabilizing the hill to prevent further landslides while clearing debris from the access road.

This is critical because: Ensuring the prevention of further landslides is crucial to allowing the safe passage of vehicles, personnel, and equipment, which will enable the restoration of access to the city and the safe evacuation of civilians.

MAIN OPERATIONS. Task 1: Stabilize the hill to prevent further landslides. Purpose 1: Enable safe debris removal from the access road. Assets performing task: Geotechnical team.
Task 2: Clear and secure the access road. Purpose 2: Restore access to the city. Assets performing task: Emergency response team equipped with heavy-duty excavation and construction equipment, disaster response units with search and rescue dogs.

AUXILIARY OPERATIONS. Task 3: Locate and rescue trapped individuals in destroyed houses and adjacent residential zones affected by the landslide. Purpose 3: Save as many lives as possible. Assets performing task: Disaster response units with search and rescue dogs, medical team for immediate on-site treatment.

END STATE:
- Assets: All teams and equipment are safely withdrawn from the area after road accessibility has been restored.
- Victims: Rescued individuals receive immediate on-site treatment from the medical team and are evacuated.
- Civilians: Civilians are informed about the stable access road and the possibility of evacuation or receiving emergency supplies.
- Terrain: The access road is clear and secured, and potential landslide areas on the hill are stabilized.

Plan Feasibility: This plan is feasible as it utilizes all available assets (emergency response team, geotechnical team, disaster response units with search and rescue dogs, and medical team) effectively and restores access to the city within the resource limitations provided.

Plan Acceptability: This plan balances the risk of the geotechnical team stabilizing the hill and emergency teams clearing and securing the road while preserving potential advantages (such as saving lives and restoring access).

Plan Suitability: This plan accomplishes the objective by prioritizing the stabilization and clearance of the access road according to the given scenario and the assets available.

Plan Distinguishability: Plan of Action 1 focuses primarily on stabilizing the hill and clearing the road to restore access, while also providing support to locate and rescue trapped individuals in destroyed houses and affected residential zones.

Total tokens: 3553 | Completion tokes: 497 | Prompt tokens: 3056

[*] User: Excellent, thank you. Please generate a sketch so I can better communicate these operations to my team members.

[*] System: Querying assistant...

[*] Plan of Action Developer Assistant: I apologize, but as a text-based AI, I am unable to create visual sketches or drawings. However, I can provide a textual description of the operations involved in the plan that you can use as a guide to create a sketch yourself.

1. Stabilize the hill:
   - Geotechnical team works to reinforce unstable areas and prevent further landslides.
   - They may use techniques like slope stabilization, retention structures, or drainage systems to secure the hill.

2. Clear and secure the access road:
   - Emergency response team equipped with heavy-duty excavation and construction equipment focuses on clearing debris from the blocked access road.
   - The disaster response units with search and rescue dogs assist in road clearance and work on rescue tasks.
 
3. Locate and rescue trapped individuals in destroyed houses and adjacent residential zones affected by the landslide:
   - Disaster response units with search and rescue dogs enter the affected houses and residential areas to search for trapped individuals.
   - The medical team provides immediate on-site treatment to rescued individuals and prepares them for evacuation.

Using this textual guide, you can create a visual sketch by drawing the hill, the access road, the affected area, and the location of different assets and teams working on each operation. Represent the teams and assets with symbols, and use arrows to indicate the direction of their actions or interventions.

Total tokens: 3850 | Completion tokes: 268 | Prompt tokens: 3582

[*] User: Thank you, that's all.

[*] System: Querying assistant...

[*] Plan of Action Developer Assistant: You're welcome! If you have any more questions or need further assistance in the future, please don't hesitate to ask. Good luck with your disaster relief operation, and stay safe!

Total tokens: 3902 | Completion tokes: 37 | Prompt tokens: 3865

[*] User: ^C

Operation interrupted by user.
\end{lstlisting}

\newpage
\subsection{Bard as the LLM backend}

\lstinputlisting{txts/exp_bard_disaster.txt}


\end{document}